# Comparative Discourse Analysis of Parallel Texts

Pim van der Eijk

Digital Equipment Corporation*
Ratelaar 38
3434 EW Nieuwegein
The Netherlands

eijk@cecamo.enet.dec.com

**Abstract**

A quantitative representation of discourse structure can be computed by measuring lexical cohesion relations among adjacent blocks of text. These representations have been proposed to deal with sub-topic text segmentation. In a parallel corpus, similar representations can be derived for versions of a text in various languages. These can be used for parallel segmentation and as an alternative measure of text-translation similarity.

## 1 Introduction

The study of large collections of texts and their translations has recently received much attention in the field of computational linguistics. In this paper, we discuss a trilingual application of earlier research on quantitative representations of the discourse structure of texts, derived from measurements of lexical cohesion. The representations are computed by measuring the similarity between vector representations of adjacent text segments, following a proposal in (Hearst, 1993). In her paper, the representations are applied to the task of segmenting long texts into sequences of discussions of subtopics, called 'tiles'. Experiments are reported that indicate that the tiles correspond rather well to human judgments on document structure.

In this paper, we apply these representations to documents of which multiple language versions are available. In a reasonably well translated parallel corpus, discourse structure seems to be a foremost property that should be preserved across translation. From the point of view of discourse analysis research, parallel

---

corpora could thus be used profitably as a resource to evaluate and compare text segmentation prototypes. From the point of view of translation research, the correlation between the vectors of similarity measurements can also be used directly as a measure of one component of text translation similarity, that can be used in addition to other measures, such as length of aligned text segments or lexical information. The techniques discussed in this paper could also be used as an alternative knowledge source for tools to align parallel documents.

This paper is structured as follows. First we discuss lexical cohesion, which is used to measure similarity of adjacent pairs of text segments. The similarity measures can be viewed as sample measurements of a 'discourse signal'. We will then briefly discuss the trilingual corpus that we used for experimentation and for evaluation, and the linguistic analysis applied to it. Different language versions of a single document will yield different discourse 'signals'. The similarity of signals can be measured by analyzing the discrete correlation of the representations.

## 2 Discourse Structure Analysis

The discourse structure of a document is analyzed by tracking patterns of semantically related elements in texts. We will first introduce some terminology, and then discuss how this structure can be computed.

### 2.1 Cohesion

In a coherent discourse, a text is not a random sequence of sentences, but rather sentences are linked by relations such as elaboration, exemplification, and cause. These relations contribute to the *coherence* of texts. There are no computational mechanisms yet to compute coherence, but it is possible to detect *cohesion*. Cohesion arises from back-references, conjunction, or lexical cohesion. Lexical cohesion is the cohesion that arises from semantic relations among words (Morris and Hirst, 1991). Lexical cohesion can be subdived in a number of classes, such as reiteration of word forms, reiteration by means of superordinates, and reiteration by means of semantically related words (either or not systematically classifiable).

For our prototype, we only detected cohesion caused by reiteration of word forms, with some provision for morphological variation (cf. section 3). Earlier research has used *Roget's Thesaurus* (Morris and Hirst, 1991) and WordNet (Hearst, 1993) as sources of semantic categorization to help detect cohesion arising from reiteration of distinct words that belong to a single semantic class. Thesaurus classes can be used instead of, or in addition to, lexical index terms. To detect reiteration of morphological variants of a word, dictionaries or morphological analysis tools are needed. For our prototype, we used lemmatizers for

English and German derived from lexical lists from Celex[1] to map unambiguous word forms to their lemma forms.

## 2.2 Computing cohesion

Cohesion relations can be used to compute the similarity of text segments. The basic approach (vector similarity measurements among adjacent pairs of text segments) is the one proposed by Marti Hearst, but some details are different, in particular the linguistic analysis described in section 3 and the choice of the digital filter. To compute similarity, segments are first analyzed as weighted vectors of index terms. Index terms are the word forms or lexemes occurring in the corpus. Term weights are computed using the idf.tf measure commonly used in vector-space approaches to information retrieval (Salton and McGill, 1983). This measure expresses that salience of terms in a segment is proportional to frequency in a segment and inversely proportional to segment frequency. A stoplist of function words can be used to restrict the attention to content words, although they have very low weights and thus only a limited influence on the similarity measure.

To compute the cohesion of a text, the similarity between the vector representations of segments is computed. The vectors can be viewed as points in a multidimensional space, where similar vectors 'point in the same direction', so that similarity can be measured using the cosine of the angle between them:

$$cos(x,y) = \frac{\sum_{t=1}^{n} w_{t,x} w_{t,y}}{\sqrt{\sum_{t=1}^{n} w_{t,x}^2} \sqrt{\sum_{t=1}^{n} w_{t,y}^2}}$$

Because terms are weighted using idf.tf, terms that are frequent in both segments, but infrequent in the document as a whole, contribute most to segment similarity. A document is then represented as a vector of cosine values corresponding to the sequence of pairs of adjacent segments, that can be plotted, yielding a wave-like figure. The waveform can be interpreted as follows: increasing values indicate continued discussion of a subtopic. Valleys mark the transition from one subtopic to another. TextTiling divides a text in regions spanning the intervals between minimal values. As discussed in (Hearst, 1993), there is a fairly good correspondence between the tiles marked by TextTiling and human judgements on text structure.

However, the measurements should be viewed as raw approximations of the discourse structure, because only a subset of cohesion relations are detected, and cohesion is only one factor contributing to coherence. The first strategy to improve on this is to improve linguistic analysis. We applied (cf. section 3) an unsophisticated morphological analysis and no semantic word class information at all. It should be noted that, for the translational applications, it is not necessary to apply a uniform analysis method to all languages. We only compare the similarity measurements, which need not be produced using identical analysis steps. This is an advantage, because in practice one often lacks comparable resources, such as a lexical list or thesaurus, for different languages.

---

[1] The Celex material is available on CD-ROM from Celex and from the LDC.

Figure 1: Effect of lowpass filtering

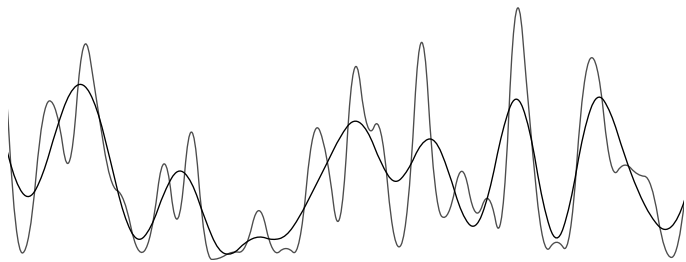

Apart from improving the analysis, a second strategy that can be applied is the use of digital filters to improve the representations. In particular, we used a low-pass filter to smooth the signal, by eliminating high-frequency components in the signal. This operation eliminates small local minima and maxima, and is needed to emphasize the general trends of the graph. To program these and other functions, we used the signal processing functions from the DXML library (DXML, 1993). In the plotted display (figure 1) the smoothed and unsmoothed representations of the English version of the UBS-corpus are displayed.

## 3  Trilingual Corpus

To evaluate the techniques, we used English, German and French versions of a banking report of the Union des Banques de Suisse (UBS) discussing developments in the Swiss economy in 1987.[2] The texts were analyzed linguistically in a number of ways. First of all, the corpus was aligned at the paragraph level. A very simple lexical analysis was subsequently applied to the three texts.

**Paragraph alignment**  In a first pass, markup was inserted in the texts to identify boundaries of 'segments' (paragraphs and headings). The texts were then semi-automatically aligned at the segment level by first matching headings, based on segment size. Headings can be distinguished from paragraphs rather easily. Small divergences (cases where two paragraphs were mapped to a single (larger) paragraph) were then easily detected, and corrected manually. The resulting corpus consisted of three parallel lists of 484 paragraphs. These are then used to generate three parallel arrays of similarity measurements, for 483 pairs of adjacent paragraphs.

---

[2]This corpus was kindly made available by Susan Armstrong, ISSCO, Geneva.

**Lexical analysis** The three language versions were analyzed lexically by removing numbers and punctiation and by converting all words to lower case. We applied a conservative type of lemmatization to the English and German versions of the document using word lists from Celex, by replacing unambiguous inflected forms by their citation forms. Ambiguous and unknown word forms were therefore left unchanged, and retained as index terms. No stoplist was used. For English, this resulted in a reduction of word form types from 4059 to 3493 for 31518 tokens. The German text contained 6233 word form types and 27019 tokens. The lower number of tokens and higher number of types is due to the productivity of compounding and spelling conventions for compounds[3] and to richer morphology. Lemmatization of unambiguous word forms reduced the number of word form types in German to 5372. Both lemmatization operations eliminate about 14% of the index terms.

The French version of the document contained 4498 word form types and 32805 tokens, and was not lemmatized for lack of a lexical list or morphological analyzer. We did evaluate a crude analysis method taking ngrams of characters as index terms instead of word forms. This approach was remarkably successful (cf. section 4), so we also applied it to the English and German texts.

## 4   Application to parallel corpora

By TextTiling, an attempt is made to capture the implicit semantic structure of a text in terms of a series of subtopics. An interesting way to evaluate and extend the use of these techniques is to apply them to a multilingual corpus.

The subtopic structure can be viewed as a property of the text which should, to some extent, be shared by a text and its translations. The paragraph similarity measurements are ultimately based on repetition of lexical material. These repetitions need not necessarily hold for the parallel segment pairs, e.g. two occurrences of a word might be translated by two synonyms, which will not be recognized as lexical cohesion in the absence of a thesaurus. In our corpus, this problem already arises by the poor morphological analysis. Another problem is noise arising from word sense ambiguity: a term might be used in distict senses, resulting in a spurious case of lexical cohesion.

Nevertheless, we hypothesized that, at the paragraph level, these discrepancies would more or less 'level out', i.e. one divergence might be offset by another convergence. This turned out to be the case, as illustrated graphically in figure 2, for part of the German and English versions, analyzed using character trigram index terms. The overall 'shape' of the curve computed is indeed largely similar. Furthermore, cases of word occurrences in a local context having distinct word senses appear to be rare in practice (Gale et al., 1992a).

---

[3] The compounding issue can be shown to be a problem for detection of word correspondences. When dealing with Dutch, which is similar to German in this respect, detection of correspondences was found to be done best at the phrase level rather than the word (form) level (van der Eijk, 1993).

Figure 2: Comparison of $de_{3gr}$ and $en_{3gr}$

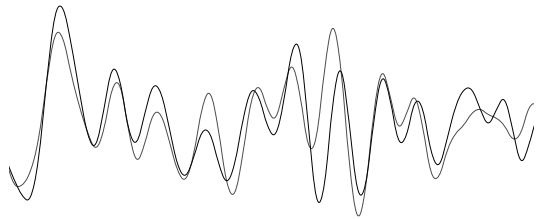

## 4.1 Measuring similarity

For each language, a vector of similarity measurements of paragraphs is generated using the method described in section 2.2. The measurements of the three versions of the UBS document can be viewed as three approximations of a single, 'underlying' discourse structure. The similarity of the three vectors of measurements is shown graphically by plotting the measurements.

To actually quantify the similarity of the paragraph similarity measurements, we computed the correlation of two arrays of measurements using a discrete summing technique. This results in an array $h$ of correlation coefficients:

$$h_j = \sum_{k=0}^{n_h-1} x_{(j+k)} y_k$$

for $j = 0, 1, 2, \ldots, n_h - 1$ and $n_h = n_x + n_y - 1$. Here, $n_h$ is the total number of points to be output from the correlation routine, and $n_x$ ($= n_y$ in our case) the number of points in the $x$ array (DXML, 1993). The values in the $h$ array can be normalized to fit in the interval $[0, 1]$ by dividing the values by the product of the norm of the $x$ and $y$ vectors. Only the first correlation coefficient is relevant, because there are no phase shifts, since the $x$ and $y$ input arrays are perfectly parallel.

Variation of analysis methods (e.g. whether or not words are lemmatized) yields slightly different paragraph similarity vectors. The correlation routine can be used to quantify the effect of using another analysis method on discourse structure, and to determine how similar various language versions of a single document are, in terms of discourse similarity. One can also turn the argument around and use correlation as a guideline to evaluate different analysis methods. If morphological analysis is hypothesized to help detect lexical cohesion, then two language versions of a document should be more strongly correlated when index terms are selected using morphological analysis.

Figure 3: Correlation of trilingual corpus

|                | $de_{nm}$ | $de_m$ | $en_{nm}$ | $en_m$ | $fr_{nm}$ | $fr_{3g}$ |
|----------------|-----------|--------|-----------|--------|-----------|-----------|
| $de_{nm}$      | 1         |        |           |        |           |           |
| $de_m$         | 0.976     | 1      |           |        |           |           |
| $en_{nm}$      | 0.86      | 0.90   | 1         |        |           |           |
| $en_m$         | 0.87      | 0.90   | 0.996     | 1      |           |           |
| $fr_{nm}$      | 0.80      | 0.81   | 0.87      | 0.88   | 1         |           |
| $fr_{3g}$      | 0.80      | 0.82   | 0.91      | 0.92   | 0.95      | 1         |

|           | $de_{3gr}$ | $en_{3gr}$ | $fr_{3gr}$ |
|-----------|------------|------------|------------|
| $de_{3gr}$| 1          |            |            |
| $en_{3gr}$| 0.97       | 1          |            |
| $fr_{3gr}$| 0.94       | 0.94       | 1          |

The correlation of the three arrays of paragraph similarity measurements, after lowpass filtering, is shown in a correlation matrix. In the matrix shown in figure 3 we have included two analyses of German, one with ($de_m$), and another without ($de_{nm}$) lemmatization, two analyses of the English, and two analyses of French, one of which is based of trigram morphology ($fr_{3g}$).

As shown, morphological analysis, and even character trigram analysis[4], resulted in a consistent, but small, improvement in measuring the similarity of the documents. The ngram analysis turned out to be superior to the analysis derived by lemmatization of unambiguous inflected word forms (cf. figure 3). Apparently, the lemmatization technique used misses a considerable number of morphological relations that can be captured with ngram analysis.

It will now be clear why the paragraph alignment phase was applied to the corpus before further analysis. Without alignment, the measurements would not be comparable, and the correlation measure would be meaningless.

## 4.2 Parallel Segmentation

Instead of computing the correlation of the representations as a measure of document similarity, it is also possible to use the representations for text segmentation as in TextTiling. Minimal values are detected and used as segment boundaries. One can then check whether representations of distinct language versions have a similar segment structure, i.e. if there is a transition from one subtopic to another in the discussion, then this transition should be detected in all three documents.

---

[4] The trigram analysis in turn improved on an analysis based on character fourgrams, fivegrams, and sixgrams (in that order).

Obviously, this is a weaker notion of similarity than discrete correlation because much information in the representations is ignored, because the paragraph similarity measurements are replaced by boolean values, viz. whether or not a gap between two paragraphs is a sub-topic boundary. Furthermore, agreement should be normalized for segment length, because distortions in segmentation are more likely to occur when segments are longer. This is the case when the representations are modified by lowpass filtering.

Parallel segmentation can be implemented in various ways. The reliability or 'strength' of a boundary can be determined by checking whether the boundary is confirmed by other language versions. If the measurements on three documents indicate a segment boundary between two paragraphs, then one will be fairly confident that there is indeed a transition to another subtopic, whereas if only one document indicates a boundary, then this is probably an incorrect measurement. We also found a number of cases of weak distortions, where two languages agree on a boundary, and the third one puts the boundary one paragraph earlier or later. An example of this is given in figure 4, where we have indicated some occurrences of segment boundaries in three language versions. These are near misses that a segmentation tool could detect and correct automatically.

Although clear correspondence between documents is an indication that the texts are related by translation, lack of correspondence (overall, or locally) can result from various reasons. One reason could be an alignment error, or a serious translation error (e.g. an untranslated section), but translation to synonyms, ambiguity, and errors in morphological analysis could cause distortions locally even when the translation is basically correct. In the specific case of the UBS corpus, some local distortions arise when there really is hardly any multi-paragraph structure at all, such as when a sequence of paragraphs is an enumeration of short overviews of economic developments in widely different sectors in the economy.

## 5  Discussion

Obviously, the quantitative techniques we have used in this paper to perform comparative discourse analysis of parallel texts are very unsophisticated, using only a subset of lexical cohesion relations, and ignoring all sub-paragraph structure. We have not evaluated the method with text units smaller than paragraphs.

Although some improvements can be obtained, esp. by applying a wider range of lexical cohesion relations, the parallel discourse analysis method discussed in this paper is best viewed as a pre-processor for other tasks. Some areas in which it can be applied are text translation alignment, translation studies, evaluation of subtopic structuring techniques, and (monolingual and multilingual) tools that use distributional information from text corpora.

Figure 4: Weak distortions (segment boundaries for $de_{3gr}$, $en_{3gr}$ and $fr_{3gr}$ )

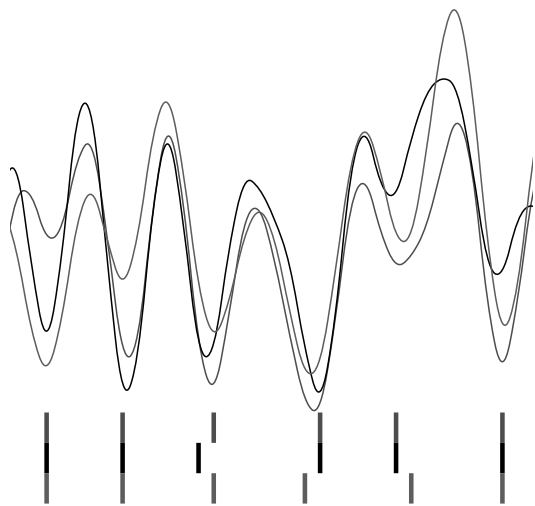

To date, systems exist that align translated documents at several levels of granularity. The first papers focussed on alignment at the sentence level (Brown et al., 1991). Recent papers have discussed alignment and correspondences at the level of words (Dagan et al., 1993) or phrases (van der Eijk, 1993). This paper has discussed how these techniques could be complemented with an algorithm to align texts at a multi-paragraph level. Earlier alignment algorithms have been remarkably successful in using only the length of parallel segments. Other algorithms have also taken lexical distribution into account (Kay and Roescheisen, 1993; Chen, 1993). Since the discourse similarity measurements have been shown to correlate strongly, this measure could also be used by alignment algorithms that can measure text-translation similarity based on more than one parameter, thus combining evidence based on segment size, lexical information, and discourse cohesion similarity. The computational overhead needed to compute cohesion similarity is very limited.

Subtopic structuring of large documents is useful for a variety of applications. Effectiveness of information retrieval on full-length documents has been shown to improve by taking advantage of document structure (Hearst and Plaunt, 1993). The availability of parallel corpora (where discourse structure is pre-served in the translation) will greatly help designing and evaluating such text segmentation systems.

Disambiguation algorithms such as (Yarowsky, 1992) that train on arbitrary-size text windows and algorithms that use lexical co-occurrence to determine semantic relatedness (Schuetze and Pedersen, 1993) might also benefit from using windows with less arbitrary boundaries. This naturally extends to similar algorithms that use distributional information in parallel corpora (Gale et al., 1992b).